\documentclass{article}

\usepackage{microtype}
\usepackage{graphicx}
\usepackage{subcaption}
\usepackage{booktabs} 

\usepackage{amsmath}
\usepackage{amssymb}
\usepackage{mathtools}
\usepackage{amsthm}

\usepackage{wrapfig}  
\usepackage{makecell}
\usepackage{multirow} 
\usepackage[utf8]{inputenc} 
\usepackage[T1]{fontenc}    
    \usepackage{url}            
\usepackage{amsfonts}       
\usepackage{nicefrac}       
\usepackage{microtype}      
\usepackage{definitions}
\usepackage{tabularx}
\usepackage[utf8]{inputenc}

\usepackage{hyperref}

\usepackage[preprint]{icml2026}

\usepackage[capitalize,noabbrev]{cleveref}

\theoremstyle{plain}

\theoremstyle{definition}

\theoremstyle{remark}

\icmltitlerunning{Context-Free Synthetic Data Mitigates Forgetting}

\begin{document}

\twocolumn[
  \icmltitle{Context-Free Synthetic Data Mitigates Forgetting}



  \icmlsetsymbol{equal}{*}

  \begin{icmlauthorlist}
    \icmlauthor{Parikshit Bansal}{sch}
    \icmlauthor{Sujay Sanghavi}{sch}
  \end{icmlauthorlist}

  \icmlaffiliation{sch}{UT Austin}

  \icmlcorrespondingauthor{Parikshit Bansal}{pbansal@utexas.edu}

  \icmlkeywords{Machine Learning, ICML}

  \vskip 0.3in
]



\printAffiliationsAndNotice{}  

\begin{abstract}
  Fine-tuning a language model often results in a degradation of its existing performance on other tasks, due to a shift in the model parameters; this phenomenon is often referred to as (catastrophic) forgetting. We are interested in mitigating this, in settings where we only have access to the model weights but no access to its training data/recipe. A natural approach is to penalize the KL divergence between the original model and the new one. Our main realization is that a simple process - which we term {\em context-free generation} - allows for an approximate unbiased estimation of this KL divergence. We show that augmenting a fine-tuning dataset with context-free generations mitigates forgetting, in two settings: {\em (a)} preserving the zero-shot performance of pretrained-only models, and {\em (b)} preserving the reasoning performance of thinking models. We show that contextual synthetic data, and even a portion of the pretraining data, are less effective. We also investigate the effect of choices like generation temperature, data ratios etc. We present our results for OLMo-1B for pretrained-only setting and R1-Distill-Llama-8B for the reasoning setting.
\end{abstract}

\section{Introduction}

It is now common practice for (so-called) ``foundation" large language models (LLMs) to be trained, with great care and at great expense, so as to be broadly performant. Specifically, these models possess very good zero-shot performance on a wide variety of tasks, including ones they may not have been specifically trained for; indeed models are now compared against each other based on how this zero-shot performance places them on multiple leaderboards. It is also common for such foundation models to be ``made public", in a very specific sense of the word: the model weights are publicly accessible and usable, but the training data, recipe etc. used to make the model are not only unavailable, but often unspecified. This means that such models can be freely used and modified, without knowledge of how they were developed. 

That being said, there are often some tasks or scenarios (e.g. involving specialized domains, or new previously unavailable data) where foundation models may not work well zero-shot. A natural and common practice in such cases is to fine-tune the model on new data aligned with these new tasks, so as to improve its performance on them. However, it is now well recognized that doing so can result in a degradation of the model's original zero-shot performance -- often, the very metrics on which it was judged to be a good model -- due to a shift in the model weights. This phenomenon is colloquially termed as ``catastrophic" {\bf forgetting}, and there are now a host of methods that attempt to mitigate forgetting; we review them in detail in our related work section. We are especially interested in methods applicable to the setting where we do not have access to the model's training data or recipe, which has been termed the {\bf data-oblivious} setting.
\begin{figure*}[t]
    \centering
    \includegraphics[width=0.7\linewidth]{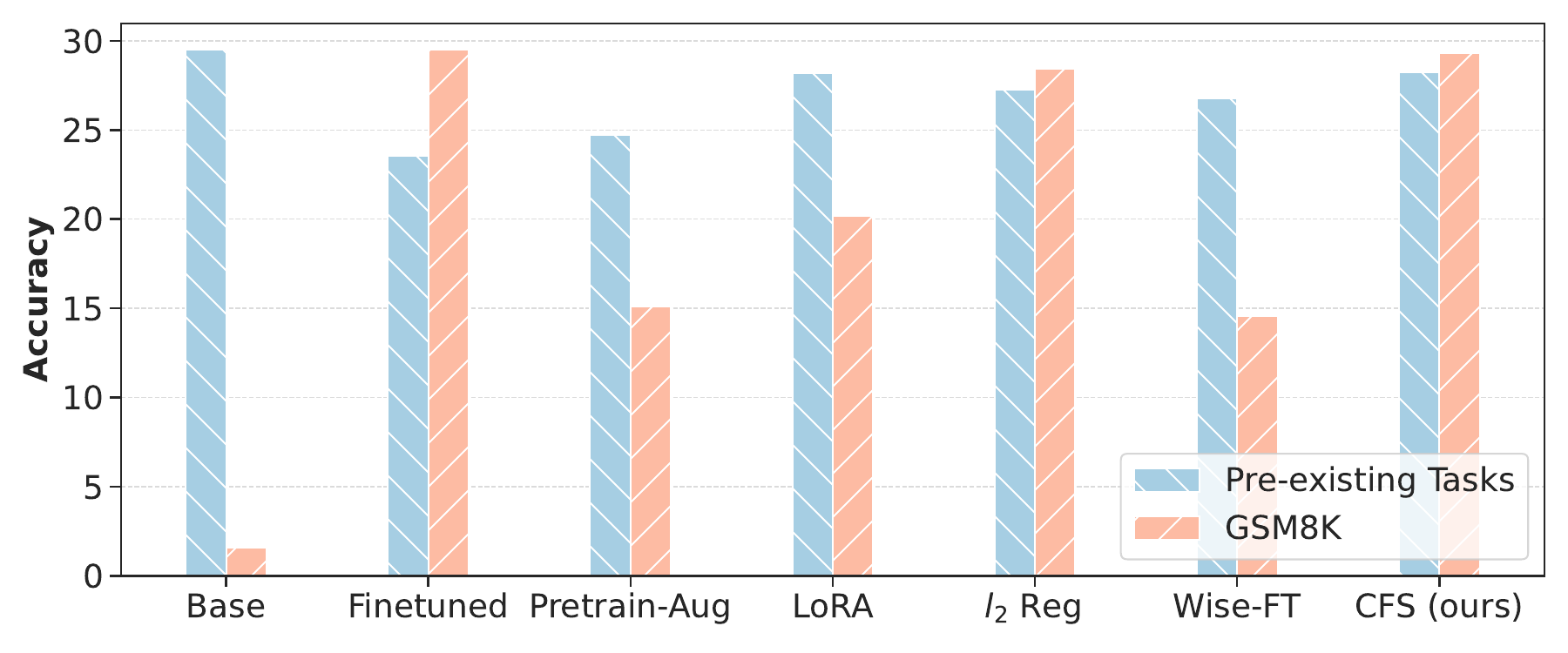}  
    \caption{We finetune Olmo-1B~\cite{groeneveld2024olmo} model on MetaMathQA~\cite{yu2023metamath} dataset with the aim of improving GSM8K accuracy while maintaining it's pre-existing (i.e., pretrained) abilities (kindly refer to Sec~\ref{subsec:pretrainingresults} for details). Our method \texttt{CFS}, augments the downstream data with context-free synthetic data (Sec~\ref{sec:context_free}) and performs better than the considered baselines. Pretrain-Aug augments MetaMathQA with pretraining data, LoRA trains a low-rank adaptation, $l_2$ regularization regularizes model towards it's initialization and Wise-FT does post-hoc model averaging of Finetuned and Base.}
    \label{fig:fig1}
\end{figure*}

Forgetting occurs because model weights shift during fine-tuning; attenuating this shift attenuates forgetting, while also possibly attenuating the new-task gains from fine-tuning. This is the realization underlying several existing methods to mitigate forgetting in the data-oblivious setting; these include adding an $\ell_2$ regularization penalty to the change in weights \cite{kirkpatrick2017overcoming,kumar2023maintaining}, deliberately using LoRA while freezing weights in the main model \cite{hu2022lora,biderman2024lora}, freezing subsets of parameters \cite{chen2024mofo,panda2406lottery}, model-averaging \cite{wortsman2021robust}, selecting easy samples during fine-tuning \cite{sanyal2025upweighting} etc. 

{\bf Our approach} starts from a simple premise: to minimize forgetting, add a penalty function that directly minimizes the shift between the resulting model and the original model. Viewing a language model as a probability distribution over sequences of tokens, a natural such penalty function would be the ``KL-divergence" between the two distributions. Of course, this is not a directly practicable idea, since there is no real way to measure/quantify this KL divergence. However, as we show below, if one could (in principle) generate an unconditional sample from the original model, one could develop an unbiased estimate of this KL divergence in a strict mathematical sense. 

However, it is not a-priori clear what it means to have an ``unconditional sample" from an LLM. Recall that inference in LLMs is typically in ``input-output" mode, i.e. outputs are produced based on a provided input context -- i.e. typical LLM inference is conditional generation. Our key realization is that having the model generate when given only just the appropriate ``begin of sentence token" but an otherwise empty input (we describe the process in detail in Section \ref{sec:context_free} for our models) serves as an effective surrogate to unconditional generation for our purpose. We term such generations {\bf context-free synthetic data (CFS)}. Under the assumption that a context-free generated string represents an unconditional sampling from the original model, an all-token pre-training-style loss on this string represents an unbiased estimate of the KL-divergence between the distributions represented by the original model and the new model, respectively.

\begin{figure*}[t]
    \centering
    \includegraphics[width=0.7\linewidth]{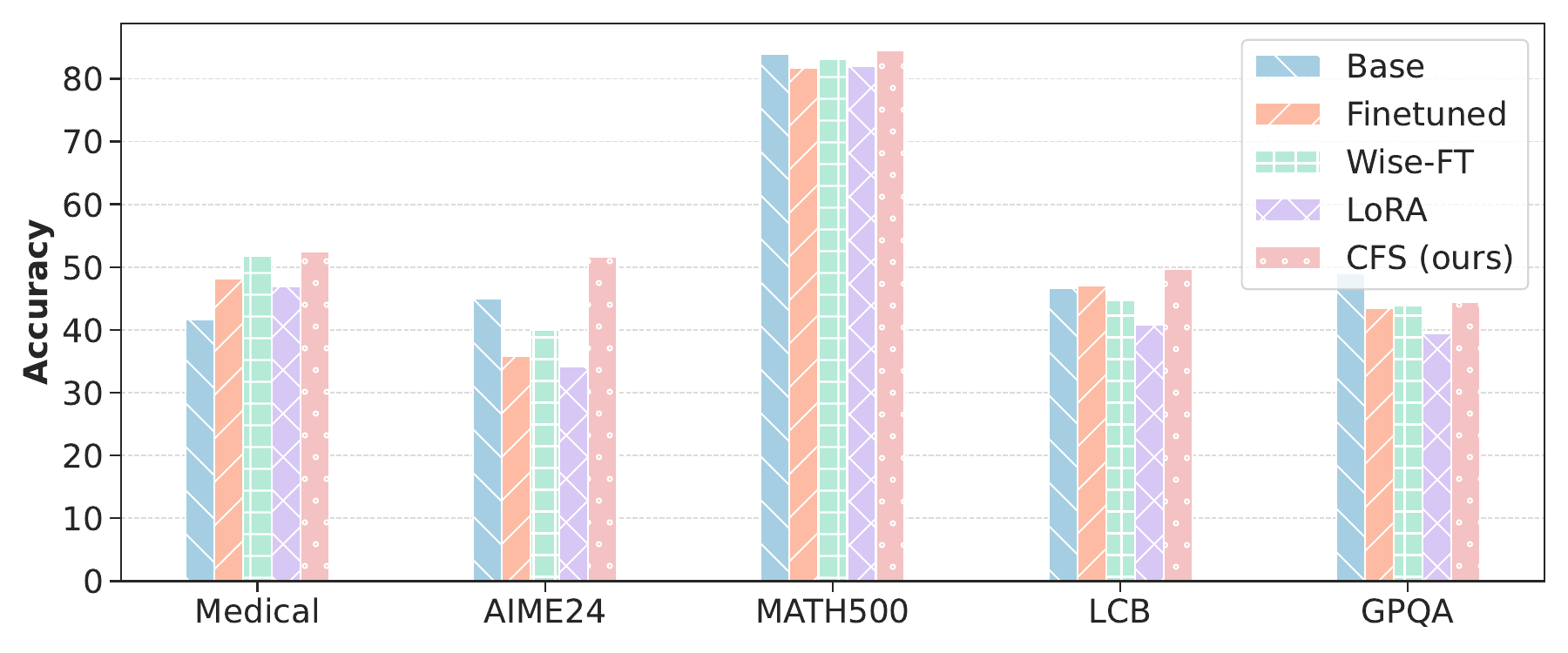}  
    \caption{We finetune R1-Distill-Llama-8B~\cite{deepseekai} model on MedReason~\cite{wu2025medreason} dataset with the aim of improving it's medical abilities, while maintaing it's reasoning performance (kindly refer to Sec~\ref{subsec:reasoningresults} for details). Our method \texttt{CFS}, augments the downstream data with context-free synthetic data (Sec~\ref{sec:context_free}) and performs better than the considered baselines, namely LoRA which trains a low-rank adaptation,  which regularizes model towards it's initialization and Wise-FT which does post-hoc model averaging of Finetuned and Base.}
    \label{fig:fig2}
\end{figure*}

Our resulting {\bf method} is straightforward: given a model whose abilities we need to not forget, and a fine-tuning dataset, first {\em (a)} generate context-free synthetic data from the model, and {\em (b)} update the model using a weighted combination of the standard fine-tuning loss on the fine-tuning dataset, and an all-token pretraining-style loss on the context-free synthetic data. Figures \ref{fig:fig1} and \ref{fig:fig2} demonstrate the forgetting problem, and also the ability of our method to mitigate it (both in absolute, and in comparison to other popular methods for the same setting).

Our work doesn't claim novelty in proposing synthetic data augmentation for mitigating forgetting. Our aim with this work is to show that our simple context-free generations from a language model are (surprisingly) better suited for mitigating forgetting (in context of LLMs) than intuitive naive choices including :  
(a) \textit{domain-specific contextual generations} : here the input context for generation are given by the input prompts of the finetuning data itself. 
(b) \textit{pretraining data} : for settings where we consider pretrained-only model with open access to it's pretraining data, our results show that context-free generations outperform augmentation with the pretraining data as well (which is equivalent to data-replay augmentation).
We highlight that work is focused on evaluating the effectiveness of different sources of data for mitigating forgetting and doesn't address practical challenges like cost of generation etc. 

Our {\bf main contributions} are:
\begin{itemize}
    \item We consider the task of mitigating catastrophic forgetting in the data oblivious setting: we are given a model to finetune on a downstream dataset but without degrading its existing capabilities, and without access to the model's original training data. 
    \item We develop a {\bf new approach}, based on viewing LLMs as probability distributions over strings. The idea behind our approach is to add a penalty term to the standard finetuning loss, where the penalty term is the KL divergence between the two distributions. We show that this KL divergence can be effectively unbiasedly estimated if one could generate unconditional samples from the LLM distributions.
    \item Based on this realization, we propose the use of {\bf context free synthetic data} generation -- basically, LLM inference from  a vacuous input context -- as a surrogate for unconditional generation. We show that this results in KL-penalization becoming our simple 2-step {\bf method:} first generate context-free synthetic data, and then fine-tune with a weighted combination of the standard fine-tuning loss on the downstream dataset and a pre-training style loss on the new synthetic data.
    \item We first demonstrate the efficacy of this method when adding a new task to a {\bf pre-trained only model}; specifically, we consider the Olmo-1B \cite{groeneveld2024olmo} model\footnote{We chose the Olmo-1B model because it's pretraining data is actually available, and we wanted to benchmark the efficacy of our data-oblivious method against classical ``replay based" continual learning approaches \cite{rolnick2019experience} that need and use pre-training data.}, which is then fine-tuned on the MetaMathQA dataset \cite{yu2023metamath} in an attempt to improve its performance on GSM8K. We show that our method works better than other benchmark data-oblivious methods like $\ell_2$ regularization, LoRA and model averaging, as well as data-aware methods like replay. See Figure \ref{fig:fig1} and Section \ref{subsec:pretrainingresults}.
    \item We then turn our attention to {\bf reasoning models}, and ask the question of whether fine-tuning degrades reasoning capacity (it does), and how to prevent this forgetting. Specifically, we investigate the effect on math reasoning performance of R1-Distill-Llama8B \cite{deepseekai} when it is fine-tuned on the medical MedReason \cite{wu2025medreason} dataset. Here again we show that our method improves on other benchmark data-oblivious methods in mitigating forgetting. See Figure \ref{fig:fig2} and Section \ref{subsec:reasoningresults} for details.
\end{itemize}

Mitigating forgetting in the data-oblivious setting is a natural and pertinent problem in the modern environment of open-weights but closed-everything-else models that are both expensive to train and need to be specialized to downstream tasks; we hope this paper adds to our understanding of this problem.

\subsection{Related work}

Catastrophic forgetting ~\cite{wang2024comprehensive,10.1145/3716629} has a rich literature, highlighting the importance of the problem. Methods to mitigate forgetting, termed as continual learning methods can broadly be classified as data-oblivious approaches (i.e., they don't assume access to prior data used to train the model) and data-dependent approaches (assume access to some subset of the data). Our focus in this paper is on the data-oblivious setting and we review relevant works in this section.

\textbf{Regularization-based approaches}
A class of well known and intuitive approaches for mitigating forgetting in data-oblivious setting constrain the learned model weights to be close to the initial model weights in a suitable metric. A simple idea is to constraint the learned weights to be close in the $\ell_2$ norm~\cite{kumar2023maintaining,kirkpatrick2017overcoming}. LoRA~\cite{hu2022lora} enforces a low rank difference between the weight matrices of the learned weights and the initial weights. ~\cite{biderman2024lora} shows that LoRA indeed mitigates forgetting, though can also hurt effective adaptation to new tasks. Another line of work doesn't constraint the problem while training, but instead post-hoc averages (or a smarter convex combination) the learned model weights with initial weights to tradeoff between learning and forgetting~\cite{lubana2021quadratic, wortsman2021robust, ilharco2023editing, lin2023mitigating, kleiman2025soupgomitigatingforgetting}. 

\textbf{Synthetic data-based approaches} A line of work in data-dependent approaches focuses on caching samples from previously seen tasks and augmenting using these samples when finetuning  for the new task~\cite{de2019episodic,rolnick2019experience}. Since the data-oblivious setting is a more realistic scenario, prior work propose using a generative model to stand-in for the previously seen data. Specifically, they jointly train both a generative model and a classifier with the generative model standing in as a proxy for previously seen tasks ~\cite{wu2018incremental,kemker2018fearnet,smith2021always,yin2020dreaming}. ~\cite{huang2024mitigating} proposes a similar setup using language model. It prompts the language model to synthesis examples similar to ones seen in the previous tasks.  

\textbf{Other data-agnostic methods } Apart from theses techniques, ~\cite{sanyal2025upweighting} explores reweighing data for mitigating forgetting. Specifically, they use the base model's own likelihood for finetuning to re-weigh samples before finetuning. They up-weigh easy samples and show that this mitigates forgetting.  \cite{chen2024mofo,panda2406lottery} also propose data-agnostic methods for continual learning which leverage gradient and other information to select a subset of parameters to update while finetuning. Similar to our work, \cite{yangmemory} mitigates forgetting by a KL divergence term between the output token probability of the base model and the learned model. We consider a more general setup where we minimize KL-divergence over the set of all strings.

While recent works on aligning language models ~\cite{shao2024deepseekmath, deepseekai} have extensively explored training on model's own responses, with some verification in loop, how model generated synthetic data can help mitigate forgetting is still under-explored. We show that particularly context-free generations are helpful for mitigating forgetting.

\begin{table*}[t]
\centering
\begin{tabular}{lll}
\toprule
\textbf{Model} & \textbf{Input Prompt} & \textbf{Examples for Context-Free Generation} \\
\midrule
Olmo-1B & \olmobos & \makecell[l]{The coronavirus pandemic has caused changes\\  in people's life as never before, with many\\ people avoiding public spaces and adhering ...}\\
\midrule
R1-Distill-Llama-8B &  \rllamabos &  \makecell[l]{Cary imprint is marked on this copy. \\ However, but cary hasn't yet been\\ assigned to the book. \textnewLine\textnewLine \underline{Wait}, let me  }\\
\bottomrule
\end{tabular}
\caption{\label{tab:bos} For each model we consider, we present the input prompt we use to generate \textit{context-free synthetic data}. Our input prompt is essentially the model's \bos i.e., beginning of sentence token. We can see that when prompted with just the model's corresponding \bos token, we are able to generate coherent samples which have high likelihood under the model and capture model's text distribution. For e.g., \textit{context-free} generations from R1-Distill-Llama-8B often contain the \wait token, enough though we don't provide any query to the model.}
\end{table*}

\section{Context-free synthetic data} \label{sec:context_free}

We now first describe the {\bf intuition} behind our method, and then provide its formal specification. 

{\bf Setup} Let us denote, as is the convention, a language model by $p_\theta$, with $\theta$ being the weights; in particular $p_\theta(x)$ denotes the probability\footnote{In particular, the standard notion of LLM probability of a sequence is the product of the probabilities the model assigns to each token in the sequence.} the model assigns to a string $x$. With this notation, {\bf pretraining} a model on a dataset $\mathcal{D}$ can be written as
\begin{equation}
\min_\theta \quad E_{x\sim \mathcal{D}} \, \left [ \, - \log p_\theta(x) \, \right ] \quad \text{starting from rand init} 
\label{eq:pretrain}
\end{equation}
Now suppose we are given a fine-tuning dataset $\mathcal{F}$ of $(x,y)$ pairs; let $p_\theta(y|x)$ denote the conditional probability of a string $y$ when $x$ is given as input context. Given a starting model $p_\theta^*$ , {\bf standard fine-tuning} on a given fine-tuning dataset $\mathcal{F}$ of $(x,y)$ pairs involves solving
\[
\min_\theta \quad E_{(x,y)\sim \mathcal{F}} \, \left [ \, - \log p_\theta(y|x) \, \right ] \quad \quad \text{starting from $\theta^*$}
\]
Note that here the maximization of the conditional probability denotes the fact that the loss is only calculated on the intended outputs $y$, while in pretraining (\ref{eq:pretrain}) we are maximizing the unconditional probability $p_\theta(x)$ itself.

In this setting, {\bf forgetting} happens because the above process results in $\theta$ making large moves away from the starting $\theta^*$ -- which then means that the resulting distribution $p_\theta$ will be far from the original $p_{\theta^*}$ even for other data unrelated to/far from $\mathcal{F}$. {\bf Conceptually} at least, a natural approach to mitigating this would be to add a penalty for how much the overall distribution shifts:
\begin{equation}
\min_\theta \quad E_{(x,y)\sim \mathcal{F}} \, \left [ \, - \log p_\theta(y|x) \, \right ] \; + \; \lambda \, \text{KL}(p_{\theta^*}\|p_\theta)
\label{eq:KL}
\end{equation}
where KL$(p_{\theta^*}\|p_\theta)$ stands for the Kullback-Liebler divergence between the original model $p_{\theta^*}$ and the new model $p_\theta$ (and $\lambda$ is a penalty parameter). Of course, the problem with this conceptual approach is that it is not clear what a term like KL$(p_{\theta^*}\|p_\theta)$ operationally means, or how to calculate it. 

Our {\bf main idea} is that we can {\em approximately estimate this KL divergence} by first generating unconditional ``context-free" synthetic samples from the existing model $p_{\theta^*}$, and then update the model via a weighted combination of the standard fine-tuning loss of  $\mathcal{F}$, and a ``pretraining style" loss (i.e. where the loss is applied to the entire string) on the synthetic samples. To see how this happens, if $\mathcal{X}$ denotes the ``set of all possible strings", we have that 
\begin{align*}
      & \text{KL}(p_{\theta^*}\|p_\theta) = \sum_{x\in \mathcal{X}} \, p_{\theta^*}(x) \log \left ( \frac{p_{\theta^*}(x)}{p_{\theta}(x)} \right ) \\
      & = E_{x\sim p_{\theta^*}} \left [ \log \left ( \frac{p_{\theta^*}(x)}{p_{\theta}(x)} \right ) \right ] \\
      & = E_{x\sim p_{\theta^*}} \left [ \log p_{\theta^*}(x) \right ] ~ + ~ E_{x\sim p_{\theta^*}} \left [ - \log p_{\theta}(x) \right ]
\end{align*}
Here the $x\sim p_{\theta^*}$ term denotes a (for now still hypothetical) sampling process for which the probability that a sample string $x$ is drawn from the set $\mathcal{X}$ is $p_{\theta^*}(x)$. 

Note now the first term in the last equation above does not depend on $\theta$; thus, minimizing KL$(p_{\theta^*}\|p_\theta)$ is equivalent to minimizing the second term. Putting this back into (\ref{eq:KL}) yields the following
\[
\min_\theta \hspace{0.5em} E_{(x,y)\sim \mathcal{F}} \left [ \, - \log p_\theta(y|x) \, \right ] \; + \; \lambda \, E_{x\sim p_{\theta^*}} \left [ - \log p_{\theta}(x) \right ]
\]
Notice that the second term above is basically a pre-training style loss like (\ref{eq:pretrain}); thus the overall loss combines standard finetuning on $\mathcal{F}$ and pretraining style loss on the new $x\sim p_{\theta^*}$ samples.

We now address the issue of what does it operationally mean to draw a sample $x\sim p_{\theta^*}$. Recall that $p_{\theta^*}$ is an autoregressive {\em generative} language model; however, the way language models are typically used is to provide an input context and sample from the conditional output. However, what the above requires is an {\em unconditional} sample, with no (or, empty) input context - we term this {\bf context-free} synthetic data. Table \ref{tab:bos} specifies how we achieve context-free generation in the two models considered in this paper: Olmo-1B, and R1-Distill-Llama-8B. 

With this in hand, {\bf our method} can be summarized as follows: Given a model $\theta^*$ (which we want to finetune without forgetting) and a finetuning dataset $\mathcal{F}$
\begin{enumerate}
    \item[{\bf (1)}] Generate context-free synthetic data $x\sim p_\theta^*$ from the model, as described above, and 
    \item[{\bf (2)}] Update the model via a weighted combination of the standard fine-tuning loss on the samples in $\mathcal{F}$ and pre-training style all-token loss on the context-free synthetic data.
\end{enumerate}

Note that our method is {\em data-oblivious}, in the sense that we only have the starting model $\theta^*$ but do not have the data used to train it. There are a few details under the hood: e.g. how many synthetic samples should one use, the temperature one should use to generate them, etc.; we study these in ablations (Sec~\ref{subsec:pretrainingresults}).

{\bf Connections to other approaches:} We now discuss connections and differences to other approaches to mitigate forgetting. We also compare against these in our experiments. \\
{\bf (1)} {\em $\ell_2$ penalization:} while our penalty term KL$(p_{\theta^*}\|p_\theta)$ penalizes shift in distributions, one could instead directly penalize a shift in the parameters themselves, i.e. have a penalty term $\|\theta - \theta^*\|^2_2$. One disadvantage of this method is that it needs to keep two sets of model weights ($\theta$ and $\theta^*$) in memory while training, while ours does not; one disadvantage of our method is that synthetic data generation may be slow because it is sequential. \\
{\bf (2)} {\em Conditional generation:} A natural synthetic data approach one may think of is to augment the fine-tuing data with condtional generation; that is, for every $(x,y)$ in $\mathcal{F}$, generate a $\widehat{y} \sim p_{\theta^*}(y|x)$ by giving the $x$ as an input context to the model $\theta^*$; and apply the standard fine-tuning loss on both the original $(x,y)$ and the new $(x,\widehat{y})$. We show that this performs quite poorly; this is because it is pushing the model in different directions for the exact same input context.\\
{\bf (3)} {\em Replay:} The idea of experience replay~\cite{rolnick2019experience}, in our context, involves retaining some portion of the past data used to train $\theta^*$, and adding it in during fine-tuning. Of course this is not data-oblivious, and is often not applicable in modern ``open weights but not open data" regimes. For one the models in this paper - Olmo-1B - the pretraining data is available, and we compare against this for that model; the other two models are ``weights only" models and hence we cannot implement replay for them. \\
{\bf (4)} {\em LoRA and weight averaging:} Fine-tuning using LoRA is seen to forget less and learn less (since the expressivity of the search space is lower than in full finetuning). Weight averaging on the other hand first does standard fine-tuning, and then averages the model weights between the new model and the original $\theta^*$. Both methods are efficient and data-oblivious, but both perform worse than our method (and, also worse than $\ell_2$ penalization)


\begin{table*}[t]
\centering
\begin{tabular}{l|l|cccc|c|c}
\toprule
\textbf{}& \textbf{}& \multicolumn{5}{c|}{\textbf{Pre-existing Tasks}}  & \\
\cmidrule(lr){2-6}
\textbf{}& \textbf{} & \textbf{Commonsense} & \textbf{MMLU} & \textbf{BB Hard} & \textbf{AGIEval} & \textbf{Avg.} & \textbf{GSM8K} \\
\midrule
\multirow{2}{*}{Standard} & \texttt{Base}  & \textbf{50.35} & \textbf{24.36} & \textbf{25.56} & {17.73} & \textbf{29.50} & 1.59 \\
& \texttt{FT} & 40.89 & 23.06 & 11.49 & 18.77 & 23.55 & \textbf{29.49} \\
\midrule
\multirow{3}{*}{Data-augmentation} & \texttt{CS}  & 42.97 & \underline{24.16} & 14.77 & 18.28 & 25.05 & 19.71 \\
& \texttt{P}  & 46.67 & 23.01 & 10.69 & 18.56 & 24.73 & 15.09 \\
& \texttt{\bf CFS} & 46.96 & 23.29 & \underline{23.19} & \underline{19.52} & \underline{28.24} & \underline{29.34} \\
\midrule
\multirow{3}{*}{Weight-Space} & \texttt{LoRA} $(r=256)$ & \underline{48.41} & {23.71} & 21.87 & 18.72 & {28.18} & 20.17 \\
& $\ell_2$ regularization  & 47.43 & 23.03 & 18.92 & \textbf{19.76} & 27.28 & 28.43 \\
&\texttt{Wise-FT}$(\alpha=0.5)$  & 45.53 & 23.10 & 19.64 & 18.85 & 26.78 & 14.56\\

\bottomrule
\end{tabular}


\caption{This table studies the pretrained-only (i.e. Olmo-1B, MetaMathQA, GSM8K) setup described in Section \ref{subsec:pretrainingresults}.  We see that standard finetuning \texttt{FT}  is better than  on \texttt{Base} the downsream task (GSM8K) but worse on the pre-existing tasks; this is forgetting. The next three rows report numbers with data-augmentation methods--including our method \texttt{CFS}. The table shows that CFS is much more effective at mitigating forgetting as compared to the other two data-augmentation-based methods \texttt{P} and \texttt{CS}.
Next we compare our method (\texttt{CFS}) to the popular weight-space approaches \texttt{LoRA}, $\ell_2$ and \texttt{Wise-FT}. Each of the baselines have hyper-parameters which we optimized over (see appendix); we report their best numbers here. We similarly ablate over the amount of context-free samples in our method (see Table~\ref{tab:cfs_ablation}) and report the best numbers here. While all these methods mitigate forgetting, our \texttt{\bf CFS} seems the most effective.
}
\label{tab:olmo}
\end{table*}

\section{Experiments} \label{sec:expts}
We consider two different experimental setups to evaluate: (a) pretrained-only models, and (b) reasoning models. In each of these setups, we first identify the following {\bf ingredients:} \\
{\bf (A)} a publicly available (i.e. ``open weights" ) model, \\
{\bf (B)} an evaluation of its zero shot performance on pre-existing tasks, and \\
{\bf (C)} a fine-tuning dataset that helps the model to improve on a new downstream task, but doing so degrades its performance on pre-existing tasks \\
In such settings, our task is to mitigate the degradation of performance on pre-existing tasks, while still benefiting the performance on downstream tasks. 

{\bf Context-free synthetic data} 
Recall that a key step in our method is context-free data generation. 
For context-free generations we prompt the model with it's \bos,i.e., beginning of sentence token. Operationally, for the two models we investigate in this paper, Table \ref{tab:bos} shows the input prompts used for the generations. We use a sampling temperature of 1.0 unless otherwise specified, and top-p of 0.95 (we ablate on the temperature hyperparameter in our experiments). We use vLLM~\cite{kwon2023efficient} for generation. 
Another choice for generating synthetic samples is contextual generations. For contextual generations, we generate model responses for each input prompt in the finetuning dataset and use the model's responses (along with the input prompts) as the augmented samples. 
For contextual-generations we use  sampling temperature of 0.6 and probability threshold (top-p) of 0.95.
For pretrained-only models, we also explore augmenting with their pretraining corpus. Specifically, for Olmo we subsample 400K rows (equal to number of samples in MetaMathQA) from it's pretraining corpus Dolma dataset~\cite{soldaini-etal-2024-dolma}. 

{\bf Methods and models} The tables in this paper study the performance of models developed using the following methods (all of which, like ours, are data-oblivious - except \texttt{P}, which in our setup is equivalent to \textit{data-replay}):

\texttt{Base}: This refers to the zero-shot performance of the base model (Olmo-1B for pre-trained-only or R1-Distill-Llama-8B for reasoning)

\texttt{FT}: This refers to the model that results when standard supervised fine-tuning is applied to the \texttt{Base} model.

\texttt{P}: We chose the Olmo-1B model because we have access to its pretraining data, which allows us to compare against the data replay approach \cite{rolnick2019experience} -- i.e. training a model on a combination of the standard fine-tuning loss on the finetuning dataset, and pretraining loss on a (random subset of) the pretraining data. Since Olmo-1B is a pretrained-only model, this approach is exactly equivalent to data-replay.

\texttt{CS}: This refers to contextual generation -- i.e. for each sample in the fine-tuning dataset, make a new sample which contains the same input context but now paired with what the old model's generated answer to that input. Fine-tune on all samples in this augmented dataset.

{\bf \texttt{CFS}}: This is {\bf our method}, based on context-free synthetic data.

\texttt{LoRA}: This refers to the model that results when LoRA~\cite{hu2022lora} is used in the standard supervised fine-tuning stage; this is based on the paper~\cite{biderman2024lora}  which shows that LoRA  can mitigate forgetting; this is thus a baseline method we compare against.

\texttt{$\ell_2$}: As described in~\cite{kumar2023maintaining,kirkpatrick2017overcoming}, this is our final baseline; it advocates adding the $\ell_2$ distance between the old model weights and the new model weights into the loss. Note that this involves needing to store two full models in memory during training, which becomes cumbersome for large models.

\texttt{Wise-FT}: Taken from \cite{wortsman2021robust}, this is another baseline method, based on model averaging. In particular, it advocates first doing standard fine-tuning, and then devising a weighted combination of $\alpha \times$ original model weights and $(1-\alpha) \times$ new model weights.

\paragraph{Training Details} We wish to investigate the effect of different augmentation sources on model performance. We use 1:1 mix of finetuning data and augmented samples, unless otherwise specified (we ablate on the mix of finetuning data and temperature in our experiments). For a fair comparison between different mix of datasets, we control for number of gradient steps, i.e. the size of dataset is inversely proportional to the number of epochs we use it for. We use the AdamW optimizer with a cosine learning rate schedule with a peak learning rate of $5e-6$ and warmup steps of $3\%$ of total training steps and train for 2-4 epochs (see Supp~\ref{appsec:training_details}). We have an effective batch size of 128 (following ~\cite{wu2025medreason,yuan2025naturalreasoning}). See Supp~\ref{appsec:training_details} for additional training details. 
We limit our investigation to just one round of finetuning as opposed to multiple rounds with multiple different tasks. Our aim is to investigate the effect of different augmentation sources in the data-mix for continual learning necessitating making this simplifying assumption.

\subsection{Pretrained-only models} \label{subsec:pretrainingresults}

\paragraph{Setup} For our first setting of pre-trained only models, we have the following ingredients: \\
{\bf (A)} We consider Olmo-1B~\cite{groeneveld2024olmo}. Olmo has publicly available  pretraining data~\cite{soldaini-etal-2024-dolma}, and is known to be not pretrained on math~\cite{groeneveld2024olmo}; hence it's 0-shot GSM8K numbers are bad. \\
{\bf (B)} Following~\citet{groeneveld2024olmo, grattafiori2024llama3herdmodels} we measure the model's performance on pre-existing tasks through eight commonsense reasoning metrics (averaged as \textit{commonsense}) along with general aggregate tasks like BigBench Hard (BB Hard), AGIEval and MMLU (see Supp~\ref{appsec:evaluation_details} for further details). \\
{\bf (C)} We train Olmo on MetaMath-QA~\cite{yu2023metamath}, and evaluate the downstream improvement by  GSM8K. Fine-tuning on this dataset helps the model improve on this benchmark, but leads to forgetting of its pre-existing performance; this sets the stage for evaluating our (and other) methods.

\begin{table}[t]
\centering
\begin{tabular}{l|cc|l|cc}
\toprule
\multicolumn{3}{c|}{\textbf{Sampling Temp.}} & \multicolumn{3}{c}{\textbf{Number of Generations}} \\
\midrule
\textbf{Model} & \textbf{Avg.} & \textbf{GSM} & \textbf{Model} & \textbf{Avg.} & \textbf{GSM} \\
\midrule
\texttt{Base}   & \textbf{29.50} & 1.59 & \texttt{Base}  & \textbf{29.50} & 1.59 \\
\texttt{FT}     & 23.55 & \textbf{29.49} & \texttt{FT}    & 23.55 & \textbf{29.49} \\
$0.6$         & 27.38 & 25.47 & 10\%           & 27.59 & 28.43 \\
$T=0.8$         & 26.89 & 26.31 & 50\%           & \underline{28.24} & \underline{29.34} \\
$T=1.0$         & \underline{27.40} & 26.00 & 100\%          & 27.38 & 26.00 \\
$T=1.2$         & 26.98 & \underline{26.54} & 200\%          & 27.81 & 22.52 \\
\bottomrule
\end{tabular}
\caption{\label{tab:cfs_ablation} In these tables, we ablate over sampling temperature (Left) and number of generated synthetic samples (Right) for the pretrain-only setup of Section \ref{subsec:pretrainingresults}.
Each table reports the model's average pre-existing tasks performance as Avg. and it's GSM8K accuracy. We see that our method is robust to choice of temperature (Left) used to sample, attenuating forgetting nonetheless. For the number of samples table (Right), different rows here correspond to different percentages (compared to downstream dataset size) of generated samples. 
\texttt{FT} can also be considered to be 0\%. We see just 10\% samples are enough for attenuating forgetting. 
}
\end{table}

\begin{table*}[t]
\centering
\small 
\begin{tabular}{l|cccc|cccc}
\toprule
 & \multicolumn{4}{c|}{\textbf{Medical}} &  \multicolumn{4}{c}{\textbf{Reasoning (Pre-existing tasks)}} \\
\cmidrule(lr){2-5} \cmidrule(lr){6-9} 
& \textbf{MedQA} & \textbf{MBOP4} & \textbf{MBOP5} & \textbf{Avg.}& {\textbf{AIME}}& {\textbf{MATH}} &  {\textbf{LCB}} & \textbf{GPQA-D}\\
\midrule
\texttt{Base}  & 47.84 & 44.81 & 32.14 & 41.60 & \underline{45.00} & \underline{84.00} & 46.62 & \textbf{48.99}\\
\texttt{FT} & {56.09} & 45.45 & {42.86} & {48.13} & 35.83 & 81.80 & \underline{47.09} & 43.43\\
\texttt{Wise-FT} $(0.5)$ & \textbf{59.78} & \textbf{51.95} & \underline{43.83} & \underline{51.85} & 40.00 & 83.20 & 44.66 & 43.94 \\ 
\texttt{LoRA} ($r=256$)  & 54.20 & 44.81 & 41.88 & 46.96 & 34.17 & 82.00 & 40.86 & 39.39 \\
\texttt{CS}  & 53.89 & {48.38} & 38.31 & 46.86 & 40.00 & 82.40 & 44.65 & 43.43\\
\texttt{\bf CFS} & \underline{59.07} & \underline{51.30} & \textbf{47.08} & \textbf{52.48} & \textbf{51.67} & \textbf{84.60} & \textbf{49.75} & \underline{44.44}\\
\bottomrule
\end{tabular}
\caption{In this table we study the reasoning setting described in Section \ref{subsec:reasoningresults} for the R1-Distill-Llama-8B model on MedReason. \texttt{FT} improves on \texttt{Base} on medical tasks, but loses its existing performance on math tasks. Among weight-space methods \texttt{Wise-FT} seems to outperform LoRA; and as opposed to the pre-trainng setting here conditional generation \texttt{CS} also works decently. Our method \texttt{CFS} outperforms all these methods. Note we could not run $\ell_2$ because our GPU constraint made the simultaneous loading two copies of the 8B model problematic.}
\label{tab:medical}
\end{table*}


{\bf Comparison against data-augmentation methods} In Table~\ref{tab:olmo}, we see that standard finetuning of Olmo on MetaMathQA leads to big increase in it's GSM8K performance, though it's pretraining abilities go down - this is forgetting. Including contextually generated data into the mix -- i.e. \texttt{CS} -- hurts GSM8K evaluation. This is intuitive as augmenting with contextually generated data brings \textit{incorrect} solutions to MetaMathQA questions into the mix. Context-free generation help the model to retain it's pretraining abilities and learn the downstream task. Surprisingly context-free generations are better than including model's pretraining data into the mix. Note that Olmo-1B is a pretrained-only model. Hence, mixing with pretraining data is equivalent to data-replay methods in continual learning. Our results suggest that in context of LLMs, context-free generations can outperform even data-replay. We hypothesize that the high variance and the noisy inherent in the pretraining data might be be the reason for the sub-optimal performance.

{\bf Comparing against weight-space methods} In Table~\ref{tab:olmo} we compare our method with relevant baselines. The main results are encapsulated in the table's caption. Importantly we see that we are better than the relevant baselines. For $l_2$ regularization we take the the regularization penalty as $1e-3$. We do a through sweep over the hyper-parameters for baselines and report the relevant numbers here. 
Specifically we sweep over the ranks of the low-rank matrices in LoRA (Table~\ref{tab:loracompletetab}), over regularization penalty for $l_2$ (Table~\ref{tab:l2completetab}) and over averaging rations $\alpha$ for Wise-FT (Table~\ref{tab:modelavgcompletetab}. \texttt{CFS} provides a better trade-off than these methods.

{\bf Ablating on number of generated samples} We study our methods performance on ablating the number of samples we generate for augmentations.
Main results are encapsulated in the caption of Table~\ref{tab:cfs_ablation}.
The number of synthetic data samples used for augmentations display the expected trend with the pre-existing task average increasing and the GSM8K performance decreasing as we increase the augmentation data. 
We would like to highlight that in the regime when the generation budget is limited (i.e., \textit{just 10\%} of the size of the finetuning data), \texttt{CFS} is still able to effectively mitigate forgetting. 

{\bf Ablating on generation parameters} 
We study our methods performance on ablating the sampling temperature in Table~\ref{tab:cfs_ablation}. We see that our method is robust to choice of temperature used to sample, attenuating forgetting nonetheless. We choose $T=1$ for rest of our experiments. We skip the ablations on other generation parameters like nucleus sampling parameter and the \texttt{CFS} generation prompt. 



\subsection{Reasoning models} \label{subsec:reasoningresults}

{\bf Setup} For the setting of reasoning models, we show that fine-tuning on a medical reasoning dataset results in degradation of the model's math capabilities. We have the following ingredients: \\
{\bf (A)} We consider R1-Distill-Llama-8B~\cite{deepseekai}. R1-Distill-Llama-8B is made by finetuning Llama-3.1-8B-Instruct~\cite{grattafiori2024llama3herdmodels} on R1 reasoning traces and is a state-of-the-art 8B reasoning model. \\
{\bf (B)} We evaluate it's existing performance on the standard reasoning benchmarks like AIME24 and MATH500, denoted by AIME and MATH resp. for math reasoning. LiveCodeBench (easy,medium,hard) with average denoted as LCB for code reasoning, and GPQA-Diamond for science reasoning. See Supp~\ref{appsec:evaluation_details} for other evaluation details. \\
{\bf (C)} We finetune R1-Distill-Llama-8B on the MedReason dataset~\cite{wu2025medreason} following the setup of~\cite{wu2025medreason}. We evaluate the model's medical abilities by MedQA~\cite{jin2021disease}, MedBulletsOp4 (MBOp4), MedBulletsOp5 (MBOp5)~\cite{wu2025medreason}. MedReason dataset has 32K samples and is constructed by converting medical question-answer pairs into reasoning steps grounded in medical knowledge-graph. This is done by structured prompting of state-of-the-art LLMs\footnote{MedReason huggingface link : \url{https://huggingface.co/datasets/UCSC-VLAA/MedReason}}.

\begin{table}[t]
\centering
\small
\begin{tabular}{l|cc}
\toprule
\textbf{Model} & \textbf{Corrections} & \textbf{Gen. Length} \\
\midrule
\texttt{Base} & 4.89 & 6737.21 \\
\texttt{FT} & 2.37 & 3933.80 \\
\texttt{Wise-FT} & 3.75 & 5239.71 \\
\texttt{LoRA} & 3.11 & 5301.22 \\
\texttt{CS} & 3.38 & 5826.27 \\
\texttt{\bf CFS} & 4.22 & 5644.58 \\
\bottomrule
\end{tabular}
\caption{For R1-Distill-Llama-8B finetuned on different augmented versions of MedReason (i.e., Table~\ref{tab:medical}), we analyze model responses for queries in MedQA, MBOP4 and MBOP5. Corrections denote the number of times the model self-corrects itself, measured by occurrences of the \wait token. Gen. Length denotes the average generation length per query. We see that while the base model has a relatively high number of self-corrections, finetuning reduces this. Context-free augmentations help retain self-correction, performing better than standard finetuning in Table~\ref{tab:medical}.}
\label{tab:selfcorrect}
\end{table}

\paragraph{Results} 
We present our main results in Table~\ref{tab:medical}. \texttt{CFS} forgets less and learns medical tasks better than the relevant baselines. Note forgetting baselines like \texttt{Wise-FT}  and our method are better than standard finetuning on MedReason. We argue that this is due to fact that evaluation benchmarks like MedQA require \textit{self-correction} ability to get good accuracy. Standard finetuning leads to a decrease in model's self-correction abilities, while improving it's medical knowledge. Finetuning while attenuating forgetting helps the model reason with the learned knowledge. We measure different model's average generations (i.e., response) length and number of self-corrections on our medical benchmarks in Table~\ref{tab:selfcorrect}. We quantify the number of self-corrections as the frequency of occurrence of \wait token per response.

We lack a comparison with $\ell_2$ regularization for this setup. This is because $\ell_2$ requires loading two models onto the GPU (the initial model and the fine-tuned model) while training. For making training these large models feasible we use DeepSpeed Zero3 parameter partition which we found is not amiable to accessing the target model parameters while training. We defer this and is on our future work.

\section{Limitations}

While our method is well-motivated, we consider two model settings: the Olmo-1B pretrained-only model (where it preserves its pretrain-model-metrics) and R1-ll-Llama-8B (where it preserves math reasoning). It would be good to see if the method works for a broader set of models and datasets.

\section*{Impact Statement}

This paper presents work whose goal is to advance the field of Machine
Learning. There are many potential societal consequences of our work, none
which we feel must be specifically highlighted here.

\nocite{langley00}

\bibliography{example_paper}

@misc{deepseekai,
      title={DeepSeek-R1: Incentivizing Reasoning Capability in LLMs via Reinforcement Learning}, 
      author={DeepSeek-AI},
      year={2025},
      eprint={2501.12948},
      archivePrefix={arXiv},
      primaryClass={cs.CL},
      url={https://arxiv.org/abs/2501.12948}, 
}

@article{yuan2025naturalreasoning,
  title={Naturalreasoning: Reasoning in the wild with 2.8 m challenging questions},
  author={Yuan, Weizhe and Yu, Jane and Jiang, Song and Padthe, Karthik and Li, Yang and Wang, Dong and Kulikov, Ilia and Cho, Kyunghyun and Tian, Yuandong and Weston, Jason E and others},
  journal={arXiv preprint arXiv:2502.13124},
  year={2025}
}

@article{wu2025medreason,
  title={MedReason: Eliciting Factual Medical Reasoning Steps in LLMs via Knowledge Graphs},
  author={Wu, Juncheng and Deng, Wenlong and Li, Xingxuan and Liu, Sheng and Mi, Taomian and Peng, Yifan and Xu, Ziyang and Liu, Yi and Cho, Hyunjin and Choi, Chang-In and others},
  journal={arXiv preprint arXiv:2504.00993},
  year={2025}
}

@article{jin2021disease,
  title={What disease does this patient have? a large-scale open domain question answering dataset from medical exams},
  author={Jin, Di and Pan, Eileen and Oufattole, Nassim and Weng, Wei-Hung and Fang, Hanyi and Szolovits, Peter},
  journal={Applied Sciences},
  volume={11},
  number={14},
  pages={6421},
  year={2021},
  publisher={MDPI}
}

@article{wang2024comprehensive,
  title={A comprehensive survey of continual learning: Theory, method and application},
  author={Wang, Liyuan and Zhang, Xingxing and Su, Hang and Zhu, Jun},
  journal={IEEE Transactions on Pattern Analysis and Machine Intelligence},
  year={2024},
  publisher={IEEE}
}

@inproceedings{kwon2023efficient,
  title={Efficient Memory Management for Large Language Model Serving with PagedAttention},
  author={Woosuk Kwon and Zhuohan Li and Siyuan Zhuang and Ying Sheng and Lianmin Zheng and Cody Hao Yu and Joseph E. Gonzalez and Hao Zhang and Ion Stoica},
  booktitle={Proceedings of the ACM SIGOPS 29th Symposium on Operating Systems Principles},
  year={2023}
}

@article{
biderman2024lora,
title={Lo{RA} Learns Less and Forgets Less},
author={Dan Biderman and Jacob Portes and Jose Javier Gonzalez Ortiz and Mansheej Paul and Philip Greengard and Connor Jennings and Daniel King and Sam Havens and Vitaliy Chiley and Jonathan Frankle and Cody Blakeney and John Patrick Cunningham},
journal={Transactions on Machine Learning Research},
issn={2835-8856},
year={2024},
url={https://openreview.net/forum?id=aloEru2qCG},
note={Featured Certification}
}

@inproceedings{
hu2022lora,
title={Lo{RA}: Low-Rank Adaptation of Large Language Models},
author={Edward J Hu and yelong shen and Phillip Wallis and Zeyuan Allen-Zhu and Yuanzhi Li and Shean Wang and Lu Wang and Weizhu Chen},
booktitle={International Conference on Learning Representations},
year={2022},
url={https://openreview.net/forum?id=nZeVKeeFYf9}
}

@article{10.1145/3716629,
author = {Zheng, Junhao and Qiu, Shengjie and Shi, Chengming and Ma, Qianli},
title = {Towards Lifelong Learning of Large Language Models: A Survey},
year = {2025},
issue_date = {August 2025},
publisher = {Association for Computing Machinery},
address = {New York, NY, USA},
volume = {57},
number = {8},
issn = {0360-0300},
url = {https://doi.org/10.1145/3716629},
doi = {10.1145/3716629},
journal = {ACM Comput. Surv.},
month = mar,
articleno = {193},
numpages = {35}
}

@article{huang2024mitigating,
  title={Mitigating catastrophic forgetting in large language models with self-synthesized rehearsal},
  author={Huang, Jianheng and Cui, Leyang and Wang, Ante and Yang, Chengyi and Liao, Xinting and Song, Linfeng and Yao, Junfeng and Su, Jinsong},
  journal={arXiv preprint arXiv:2403.01244},
  year={2024}
}

@article{de2019episodic,
  title={Episodic memory in lifelong language learning},
  author={de Masson D'Autume, Cyprien and Ruder, Sebastian and Kong, Lingpeng and Yogatama, Dani},
  journal={Advances in Neural Information Processing Systems},
  volume={32},
  year={2019}
}

@article{groeneveld2024olmo,
  title={Olmo: Accelerating the science of language models},
  author={Groeneveld, Dirk and Beltagy, Iz and Walsh, Pete and Bhagia, Akshita and Kinney, Rodney and Tafjord, Oyvind and Jha, Ananya Harsh and Ivison, Hamish and Magnusson, Ian and Wang, Yizhong and others},
  journal={arXiv preprint arXiv:2402.00838},
  year={2024}
}

@article{yu2023metamath,
  title={Metamath: Bootstrap your own mathematical questions for large language models},
  author={Yu, Longhui and Jiang, Weisen and Shi, Han and Yu, Jincheng and Liu, Zhengying and Zhang, Yu and Kwok, James T and Li, Zhenguo and Weller, Adrian and Liu, Weiyang},
  journal={arXiv preprint arXiv:2309.12284},
  year={2023}
}

@inproceedings{rolnick2019experience,
 author = {Rolnick, David and Ahuja, Arun and Schwarz, Jonathan and Lillicrap, Timothy and Wayne, Gregory},
 booktitle = {Advances in Neural Information Processing Systems},
 editor = {H. Wallach and H. Larochelle and A. Beygelzimer and F. d\textquotesingle Alch\'{e}-Buc and E. Fox and R. Garnett},
 pages = {},
 publisher = {Curran Associates, Inc.},
 title = {Experience Replay for Continual Learning},
 url = {https://proceedings.neurips.cc/paper_files/paper/2019/file/fa7cdfad1a5aaf8370ebeda47a1ff1c3-Paper.pdf},
 volume = {32},
 year = {2019}
}

@misc{grattafiori2024llama3herdmodels,
      title={The Llama 3 Herd of Models}, 
      author={Aaron Grattafiori et al. },
      year={2024},
      eprint={2407.21783},
      archivePrefix={arXiv},
      primaryClass={cs.AI},
      url={https://arxiv.org/abs/2407.21783}, 
}

@article{chen2024mofo,
  title={MoFO: Momentum-Filtered Optimizer for Mitigating Forgetting in LLM Fine-Tuning},
  author={Chen, Yupeng and Wang, Senmiao and Lin, Zhihang and Qin, Zeyu and Zhang, Yushun and Ding, Tian and Sun, Ruoyu},
  journal={arXiv preprint arXiv:2407.20999},
  year={2024}
}

@article{wortsman2021robust,
  title={Robust fine-tuning of zero-shot models},
  author={Wortsman, Mitchell and Ilharco, Gabriel and Kim, Jong Wook and Li, Mike and Kornblith, Simon and Roelofs, Rebecca and Gontijo-Lopes, Raphael and Hajishirzi, Hannaneh and Farhadi, Ali and Namkoong, Hongseok and Schmidt, Ludwig},
  journal={arXiv preprint arXiv:2109.01903},
  note={\url{https://arxiv.org/abs/2109.01903}},
  year={2021}
}

@inproceedings{kemker2018fearnet,
title={FearNet: Brain-Inspired Model for Incremental Learning},
author={Ronald Kemker and Christopher Kanan},
booktitle={International Conference on Learning Representations},
year={2018},
url={https://openreview.net/forum?id=SJ1Xmf-Rb},
}

@article{wu2018incremental,
  title={Incremental Classifier Learning with Generative Adversarial Networks},
  author={Yue Wu and Yinpeng Chen and Lijuan Wang and Yuancheng Ye and Zicheng Liu and Yandong Guo and Zhengyou Zhang and Yun Raymond Fu},
  journal={ArXiv},
  year={2018},
  volume={abs/1802.00853},
  url={https://api.semanticscholar.org/CorpusID:3652214}
}

@article{panda2406lottery,
  title={Lottery ticket adaptation: Mitigating destructive interference in llms, 2024},
  author={Panda, Ashwinee and Isik, Berivan and Qi, Xiangyu and Koyejo, Sanmi and Weissman, Tsachy and Mittal, Prateek},
    year={2024},
  journal={https://arxiv. org/abs/2406.16797}
}

@inproceedings{ilharco2023editing,
title={Editing models with task arithmetic},
author={Gabriel Ilharco and Marco Tulio Ribeiro and Mitchell Wortsman and Ludwig Schmidt and Hannaneh Hajishirzi and Ali Farhadi},
booktitle={The Eleventh International Conference on Learning Representations },
year={2023},
url={https://openreview.net/forum?id=6t0Kwf8-jrj}
}

@misc{kleiman2025soupgomitigatingforgetting,
      title={Soup to go: mitigating forgetting during continual learning with model averaging}, 
      author={Anat Kleiman and Gintare Karolina Dziugaite and Jonathan Frankle and Sham Kakade and Mansheej Paul},
      year={2025},
      eprint={2501.05559},
      archivePrefix={arXiv},
      primaryClass={cs.LG},
      url={https://arxiv.org/abs/2501.05559}, 
}

@article{lubana2021quadratic,
  title     = {How do Quadratic Regularizers Prevent Catastrophic Forgetting: The Role of Interpolation},
  author    = {Ekdeep Singh Lubana and Puja Trivedi and Danai Koutra and R. Dick},
  journal   = {COLLAS},
  year      = {2021},
  bibSource = {Semantic Scholar https://www.semanticscholar.org/paper/90a0bea2b19957e9f9d1c920b3f5885ce2323d69}
}

@article{lin2023mitigating,
  title   = {Mitigating the Alignment Tax of RLHF},
  author  = {Lin, Yong and Lin, Hangyu and Xiong, Wei and Diao, Shizhe and Liu, Jianmeng and Zhang, Jipeng and Pan, Rui and Wang, Haoxiang and Hu, Wenbin and Zhang, Hanning and others},
  journal = {CoRR},
  year    = {2023}
}

@inproceedings{soldaini-etal-2024-dolma,
    title = "Dolma: an Open Corpus of Three Trillion Tokens for Language Model Pretraining Research",
    author = "Soldaini, Luca  and
      Kinney, Rodney  and
      Bhagia, Akshita  and
      Schwenk, Dustin  and
      Atkinson, David  and
      Authur, Russell  and
      Bogin, Ben  and
      Chandu, Khyathi  and
      Dumas, Jennifer  and
      Elazar, Yanai  and
      Hofmann, Valentin  and
      Jha, Ananya  and
      Kumar, Sachin  and
      Lucy, Li  and
      Lyu, Xinxi  and
      Lambert, Nathan  and
      Magnusson, Ian  and
      Morrison, Jacob  and
      Muennighoff, Niklas  and
      Naik, Aakanksha  and
      Nam, Crystal  and
      Peters, Matthew  and
      Ravichander, Abhilasha  and
      Richardson, Kyle  and
      Shen, Zejiang  and
      Strubell, Emma  and
      Subramani, Nishant  and
      Tafjord, Oyvind  and
      Walsh, Evan  and
      Zettlemoyer, Luke  and
      Smith, Noah  and
      Hajishirzi, Hannaneh  and
      Beltagy, Iz  and
      Groeneveld, Dirk  and
      Dodge, Jesse  and
      Lo, Kyle",
    editor = "Ku, Lun-Wei  and
      Martins, Andre  and
      Srikumar, Vivek",
    booktitle = "Proceedings of the 62nd Annual Meeting of the Association for Computational Linguistics (Volume 1: Long Papers)",
    month = aug,
    year = "2024",
    address = "Bangkok, Thailand",
    publisher = "Association for Computational Linguistics",
    url = "https://aclanthology.org/2024.acl-long.840/",
    doi = "10.18653/v1/2024.acl-long.840",
    pages = "15725--15788"
}

@article{sanyal2025upweighting,
  title={Upweighting Easy Samples in Fine-Tuning Mitigates Forgetting},
  author={Sanyal, Sunny and Prairie, Hayden and Das, Rudrajit and Kavis, Ali and Sanghavi, Sujay},
  journal={arXiv preprint arXiv:2502.02797},
  year={2025}
}

@article{shao2024deepseekmath,
  title={Deepseekmath: Pushing the limits of mathematical reasoning in open language models},
  author={Shao, Zhihong and Wang, Peiyi and Zhu, Qihao and Xu, Runxin and Song, Junxiao and Bi, Xiao and Zhang, Haowei and Zhang, Mingchuan and Li, YK and Wu, Y and others},
  journal={arXiv preprint arXiv:2402.03300},
  year={2024}
}

@inproceedings{smith2021always,
  title={Always be dreaming: A new approach for data-free class-incremental learning},
  author={Smith, James and Hsu, Yen-Chang and Balloch, Jonathan and Shen, Yilin and Jin, Hongxia and Kira, Zsolt},
  booktitle={Proceedings of the IEEE/CVF international conference on computer vision},
  pages={9374--9384},
  year={2021}
}

@inproceedings{yin2020dreaming,
  title={Dreaming to distill: Data-free knowledge transfer via deepinversion},
  author={Yin, Hongxu and Molchanov, Pavlo and Alvarez, Jose M and Li, Zhizhong and Mallya, Arun and Hoiem, Derek and Jha, Niraj K and Kautz, Jan},
  booktitle={Proceedings of the IEEE/CVF conference on computer vision and pattern recognition},
  pages={8715--8724},
  year={2020}
}

@article{kumar2023maintaining,
  title={Maintaining plasticity in continual learning via regenerative regularization},
  author={Kumar, Saurabh and Marklund, Henrik and Van Roy, Benjamin},
  journal={arXiv preprint arXiv:2308.11958},
  year={2023}
}

@article{yangmemory,
  title={Memory retaining finetuning via distillation},
  author={Yang, Zitong and Zhang, Aonan and Wiseman, Sam and Kong, Xiang and Ye, Ke and Yin, Dong}
}

@article{kirkpatrick2017overcoming,
  title={Overcoming catastrophic forgetting in neural networks},
  author={Kirkpatrick, James and Pascanu, Razvan and Rabinowitz, Neil and Veness, Joel and Desjardins, Guillaume and Rusu, Andrei A and Milan, Kieran and Quan, John and Ramalho, Tiago and Grabska-Barwinska, Agnieszka and others},
  journal={Proceedings of the national academy of sciences},
  volume={114},
  number={13},
  pages={3521--3526},
  year={2017},
  publisher={National Academy of Sciences}
}
\bibliographystyle{icml2026}

\newpage
\appendix
\onecolumn

\section{Compute details}

Each training run is done on a single GH200 GPU of size 96GB. Each individual run in our paper takes around 8 hours for training and about 1-2 hours for evaluation. 

\section{Training details}\label{appsec:training_details}
We use the 8bit AdamW (from bitsandbytes). 
We finetune Olmo on MetaMathQA for 2 epochs. R1-Distill-Llama-8B is finetuned on Medreason for 4 corresponding to approx. 1000 gradient steps. We limit the training sequence to be of length 1024 for R1-Distill-Llama-8B and of length 512 for Olmo. 

For MedReason we format the reasoning traces inside a \texttt{<think>}\ldots \think block, which is preceded by the question in MedReason and followed by the answer. This follows R1-Distill's general reasoning data format and hence we use this. We don't use a system prompt either for training or evaluation.

For MetaMathQA we format question as "Question : \{question\}, Answer : \{answer\}". 

\section{Evaluation details} \label{appsec:evaluation_details}
Olmo evaluation details : 
\begin{itemize}
    \item We use LMEval\footnote{\url{https://github.com/EleutherAI/lm-evaluation-harness}} for evaluating pretraining abilities.  
    \item \textbf{Commonsense reasoning datasets} : Following~\cite{groeneveld2024olmo} we average the performance on the following datasets for commonsense reasoning : 1. ARC-challenge, 2. ARC-easy, 3. Boolq, 4. Hellaswag, 5. Openbookqa, 6. Piqa, 7. Siqa, 9. Winogrande
    \item We borrow the setup from \cite{grattafiori2024llama3herdmodels} and report numbers considered in their "general language tasks" table for pretrained models. Since Olmo is not pretrained on math and code we don't report those numbers and we found reading comprehension numbers for the models to be unreliable. 
    \item We here report the exact LMEval metrics we evaluate : For GSM8K we evaluate it's standard 5 shot performance i.e., we report \texttt{gsm8k/exact\textunderscore match,strict-match} , for AGIEval we consider only the english subset i.e. \texttt{agieval\textunderscore{en}/acc}. For BigBenchHard we use \texttt{bbh/exact\textunderscore match,get-answer}  and for MMLU we use \texttt{mmlu/acc}. 
\end{itemize}

R1-Distill-Llama-8B evaluation details : 
\begin{itemize}
    \item We uses two codebase for our evaluation : SkyThoughts\footnote{\url{https://github.com/NovaSky-AI/SkyThought}} for reasoning tasks i.e., AIME24, MATH500, LiveCodeBench and GPQA-Diamond. We use MedReason\footnote{\url{https://github.com/UCSC-VLAA/MedReason}} for medical benchmarking.
    \item We use the standard sampling params (from ~\cite{deepseekai}) for evaluating reasoning performance : temperature = 0.6, and probability threshold of 0.95. We generate for max length of 32768. We use vLLM for generating responses and inherit the reproducibility issues in it's generation. We also append a \texttt{<think>} token to the query whenever we are generating a response. 
    \item For evaluating medical benchmarks we use the specified hyperparameters in MedReason, i.e. max generation length of 2048 tokens, and temperature of 0.6 and probability threshold of 0.95. 
    \item For {AIME24} we sample 4 different solutions for each prompt making the effective test set size equal to 120. For other datasets we only use only one sample per prompt {MATH, GPQA-Diamond}. 
\begin{table}[h]
    \centering
    \begin{tabular}{c|c}
        Dataset & Effective Size \\
\hline
        AIME & 120 \\
        MATH500 & 500 \\
        GPQA-Diamond & 198 \\
        LiveCodeBench v2 & 511 \\
        MedQA & 1273 \\
        MedBulletsOp4 & 308 \\
        MedBulletsOp5 & 308 \\
    \end{tabular}
    \caption{Evaluation dataset sizes}
    \label{tab:eval_dataset_sizes}
\end{table}
    \item Our reported numbers are worse than Deepseek numbers : SkyThoughts uses regex parsing for verifying model answers and hence is under-reports the model's accuracy. Gold standard evaluation uses light models OpenAI-o1-mini to comparing the answers, which we omit for computational and financial considerations.
\end{itemize}

\












\begin{table}[h]
\centering
\small 
\begin{tabular}{l|cccc|c|c}
\toprule
\textbf{}& \multicolumn{5}{c|}{\textbf{Pretraining Abilities}}  & \\
\cmidrule(lr){2-6}
\textbf{} & \textbf{Commonsense} & \textbf{MMLU} & \textbf{BB Hard} & \textbf{AGIEval} & \textbf{Avg.} & \textbf{GSM8K} \\
\midrule
rank=64 & 47.44 & 23.68 & 21.07 & 18.46 & 27.66 & 13.04 \\
rank=128 & 47.83 & 24.33 & 21.33 & 18.59 & 28.02 & 17.66 \\
rank=256 & 48.41 & 23.71 & 21.87 & 18.72 & 28.18 & 20.17 \\
rank=64+CFS\textsuperscript{*} & 48.94 & 24.45 & 22.49 & 18.28 & 28.54 & 8.42 \\
rank=128+CFS\textsuperscript{*} & 48.91 & 23.76 & 23.59 & 18.22 & 28.62 & 9.63 \\
rank=256+CFS\textsuperscript{*} & 49.03 & 24.16 & 23.68 & 18.61 & 28.87 & 11.83 \\
 \bottomrule
\end{tabular}
\caption{LoRA hyperparameter sweep for Olmo. We also include numbers of combining CFS with LoRA.}
\label{tab:loracompletetab}
\end{table}

\begin{table}[h]
\centering
\small 
\begin{tabular}{l|cccc|cccc}
\toprule
 & \multicolumn{4}{c|}{\textbf{Medical}} &  \multicolumn{4}{c}{\textbf{Reasoning}} \\
\cmidrule(lr){2-5} \cmidrule(lr){6-9} 
& \textbf{MedQA} & \textbf{MBOP4} & \textbf{MBOP5} & \textbf{Avg.}& {\textbf{AIME}}& {\textbf{MATH}} &  {\textbf{LCB}} & \textbf{GPQA-D}\\
\midrule
$r=256$  & 54.20 & 44.81 & 41.88 & 46.96 & 34.17 & 82.00 & 40.86 & 39.39 \\
$r=512$  & 52.47 & 43.51 & 38.64 & 44.87 & 32.50 & 82.60 & 44.96 & 43.94 \\ 
 \bottomrule
\end{tabular}
\caption{\label{tab:loracompletetabreasoning} LoRA ranks tried for R1-Distill-Llama-8B. Since LoRA proves to be less expressive for our setup, we only experiment with high values of $r$.}
\end{table}

\begin{table}[h]
\centering
\small 
\begin{tabular}{l|cccc|c|c}
\toprule
\textbf{}& \multicolumn{5}{c|}{\textbf{Pretraining Abilities}}  & \\
\cmidrule(lr){2-6}
\textbf{} & \textbf{Commonsense} & \textbf{MMLU} & \textbf{BB Hard} & \textbf{AGIEval} & \textbf{Avg.} & \textbf{GSM8K} \\
\midrule
$1e-1$ & 47.81 & 23.41 & 20.87 & 17.86 & 27.49 & 6.90 \\
$1e-2$ & 47.64 & 23.49 & 21.59 & 18.41 & 27.78 & 18.57 \\
$1e-3$  & 47.43 & 23.03 & 18.92 & 19.76 & 27.28 & 28.43 \\
 \bottomrule
\end{tabular}
\caption{$\ell_2$ regularization penalty sweep for training Olmo on MetaMathQA.}
\label{tab:l2completetab}
\end{table}

\section{Finetuning dataset details}
\paragraph{MedReason} is a medical reasoning dataset designed for explainable medical problem-solving in large language models (LLMs). It utilizes a structured medical knowledge graph (KG) to convert clinical QA pairs into logical chains of reasoning, which trace connections from question elements to answers via relevant KG entities. Each path is validated for consistency with clinical logic and evidence-based medicine. They consider medical questions from 7 medical datasets, resulting in a dataset of 32,682 question-answer pairs.
\paragraph{MetaMathQA} is a large mathematical problem solving dataset. Given a meta-question, a question in train set of GSM8K, it generates a series of variants of the question. Specifically, they perform three types
of question bootstrapping. They also performa answer augmentation, leading to the 400K sample MetaMathQA dataset. MetaMathQA focusing on elementary mathematical problem-solving

\section{Complete tables}

\subsection{LoRA Complete Table}
See Table~\ref{tab:loracompletetab} for LoRA rank sweep for Olmo and Table~\ref{tab:loracompletetabreasoning} for LoRA rank sweep for R1-Distill-Llama-8B. Across both tables we see that LoRA is not able to perform at par with finetuning on downstream evaluation. We keep the the alpha parameter in LoRA equal to it's rank. 
\subsection{L2 Complete Table}
See Table~\ref{tab:l2completetab} for $\ell_2$ hyperparameter sweep for Olmo. $1e-1$ proves to be too strong of a regularization, we take $1e-3$ as having the best of both worlds. 
\subsection{ModelAvg Complete Table}
See Table~\ref{tab:modelavgcompletetab} for Model averaging results for pretrained models and Table~\ref{tab:modelavgcompletetabreasoning} for model averaging for reasoning models. For computational reasons, we evaluate less number of averaging factors for reasoning models, but the results are informative nonetheless. For Table~\ref{tab:modelavgcompletetabreasoning} We can see that as we go from a high value of $\alpha$ to a lower value, i.e. from the base model to finetuned model, we see a U-shape in the downstream performance, instead of the expected straight line we see in Table~\ref{tab:modelavgcompletetab}. We believe this is due to reasons discussed in Sec~\ref{subsec:reasoningresults}, specifically, having some base model capabilities like self-reflections and correction help the model be more accurate on the downstream evaluation.

\begin{table}[h]
\centering
\small 
\begin{tabular}{l|cccc|c|c}
\toprule
\textbf{}& \multicolumn{5}{c|}{\textbf{Pretraining Abilities}}  & \\
\cmidrule(lr){2-6}
\textbf{} & \textbf{Commonsense} & \textbf{MMLU} & \textbf{BB Hard} & \textbf{AGIEval} & \textbf{Avg.} & \textbf{GSM8K} \\
\midrule
$\alpha=0.1$ & 41.63 & 23.07 & 14.27 & 19.18 & 24.54 & 28.73\\
$\alpha=0.2$  & 42.55 & 23.10 & 15.93 & 19.29 & 25.22 & 28.58\\
$\alpha=0.3$  & 43.26 & 23.12 & 17.32 & 19.24 & 25.73 & 25.70\\
$\alpha=0.4$  & 44.37 & 23.06 & 19.15 & 19.13 & 26.43 & 20.39\\
$\alpha=0.5$  & 45.53 & 23.10 & 19.64 & 18.85 & 26.78 & 14.56\\
$\alpha=0.6$  & 46.62 & 23.19 & 21.24 & 18.33 & 27.34 & 8.04\\
$\alpha=0.7$  & 47.66 & 23.90 & 23.36 & 17.81 & 28.18 & 3.34\\
$\alpha=0.8$  & 48.62 & 24.36 & 24.37 & 17.37 & 28.68 & 2.12\\
$\alpha=0.9$  & 49.54 & 24.40 & 25.14 & 17.21 & 29.07 & 2.20\\
\bottomrule
\end{tabular}
\caption{Model Averaging results for Olmo finetuned on MetaMathQA.
}
\label{tab:modelavgcompletetab}
\end{table}

\begin{table}[h]
\centering
\small 
\begin{tabular}{l|cccc|cccc}
\toprule
 & \multicolumn{4}{c|}{\textbf{Medical}} &  \multicolumn{4}{c}{\textbf{Reasoning}} \\
\cmidrule(lr){2-5} \cmidrule(lr){6-9} 
& \textbf{MedQA} & \textbf{MBOP4} & \textbf{MBOP5} & \textbf{Avg.}& {\textbf{AIME}}& {\textbf{MATH}} &  {\textbf{LCB}} & \textbf{GPQA-D}\\
\midrule
$\alpha=0.1$ & 58.99 & 49.68 & 46.43 & 51.70 & 40.00 & 80.20 & 46.22 & 39.39 \\ 
$\alpha=0.3$ & 58.60 & 52.27 & 47.08 & 52.65 & 49.17 & 81.40 & 46.05 & 43.94 \\
$\alpha=0.5$ & 59.78 & 51.95 & 43.83 & 51.85 & 40.00 & 83.20 & 44.66 & 43.94 \\ 
$\alpha=0.7$ & 55.22 & 52.27 & 39.61 & 49.04 & 45.83 & 86.00 & 48.20 & 45.96 \\ 
$\alpha=0.9$ & 52.55 & 45.45 & 37.66 & 45.22 & 48.33 & 85.40 & 47.99 & 50.00 \\ 
\bottomrule
\end{tabular}
\caption{\label{tab:modelavgcompletetabreasoning} Model Averaging results for R1-Distill trained on MedReason dataset.}
\end{table}


\end{document}